\documentclass[manuscript,review=false]{acmart}

\usepackage{subcaption}
\usepackage{pifont}
\newcommand{\cmark}{\ding{51}}%
\newcommand{\xmark}{\ding{55}}%

\AtBeginDocument{%
  \providecommand\BibTeX{{%
    \normalfont B\kern-0.5em{\scshape i\kern-0.25em b}\kern-0.8em\TeX}}}

\setcopyright{acmcopyright}
\copyrightyear{2018}
\acmYear{2018}
\acmDOI{XXXXXXX.XXXXXXX}

\acmConference[Conference acronym 'XX]{Make sure to enter the correct
  conference title from your rights confirmation emai}{June 03--05,
  2018}{Woodstock, NY}
\acmBooktitle{Woodstock '18: ACM Symposium on Neural Gaze Detection,
 June 03--05, 2018, Woodstock, NY}
\acmPrice{15.00}
\acmISBN{978-1-4503-XXXX-X/18/06}

\begin{document}

\title{Feedback Driven Multi Stereo Vision System for Real-Time Event Analysis}

\author{Mohamed Benkedadra}
\email{mohamed.benkedadra@umons.ac.be}
\orcid{0000-0002-2590-8344}
\affiliation{%
  \institution{University of Mons}
  \city{Mons}
  \state{Hainaut}
  \country{Belgium}
  \postcode{7000}
}

\author{Matei Mancas}
\orcid{0000-0002-6539-6996}
\affiliation{%
  \institution{University of Mons}
  \city{Mons}
  \state{Hainaut}
  \country{Belgium}
  \postcode{7000}
}
\email{matei.mancas@umons.ac.be}

\author{Sidi Ahmed Mahmoudi}
\orcid{0000-0002-1530-9524}
\affiliation{%
  \institution{University of Mons}
  \city{Mons}
  \state{Hainaut}
  \country{Belgium}
  \postcode{7000}
}

\renewcommand{\shortauthors}{Benkedadra, et al.}

\begin{abstract}
2D cameras are often used in interactive systems. Other systems like gaming consoles provide more powerful 3D cameras for short range depth sensing. Overall, these cameras are not reliable in large, complex environments. In this work, we propose a 3D stereo vision based pipeline for interactive systems, that is able to handle both ordinary and sensitive applications, through robust scene understanding. We explore the fusion of multiple 3D cameras to do full scene reconstruction, which allows for preforming a wide range of tasks, like event recognition, subject tracking, and notification. Using possible feedback approaches, the system can receive data from the subjects present in the environment, to learn to make better decisions, or to adapt to completely new environments. Throughout the paper, we introduce the pipeline and explain our preliminary experimentation and results. Finally, we draw the roadmap for the next steps that need to be taken, in order to get this pipeline into production.
\end{abstract}





\begin{teaserfigure}
  \centering
  \includegraphics[width=\linewidth]{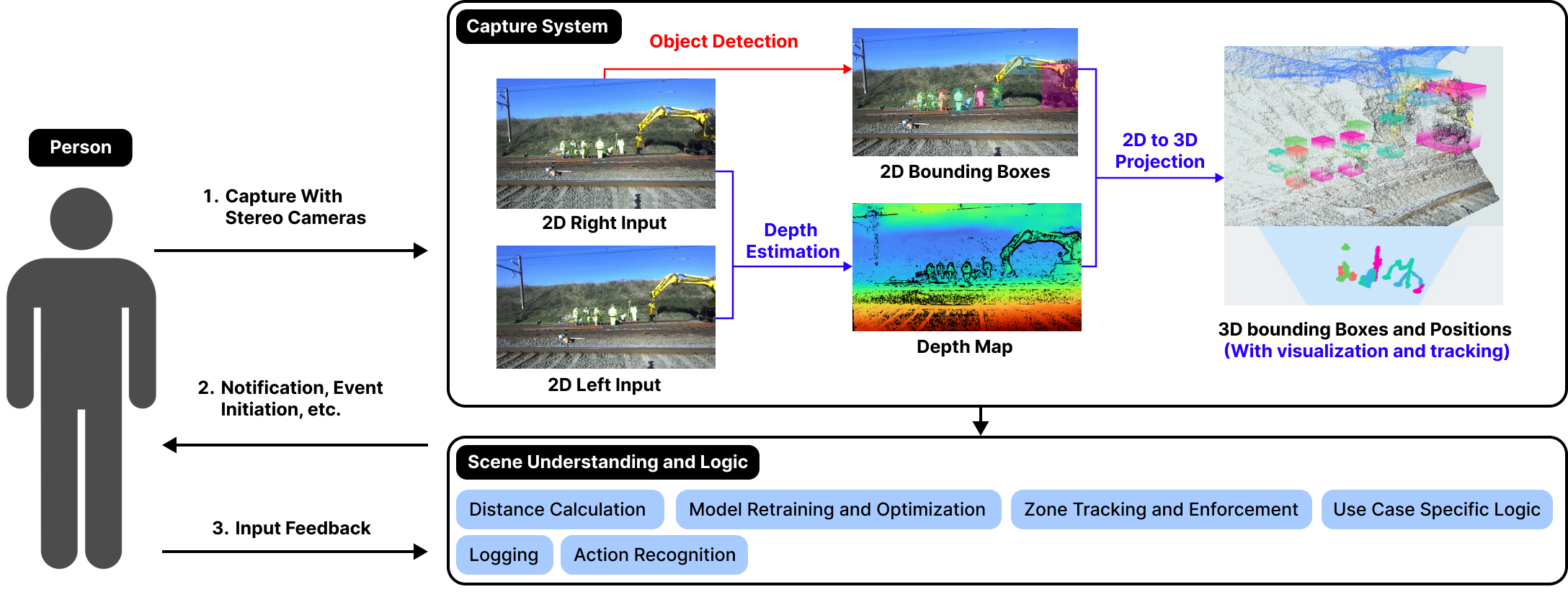}
  \caption{Proposed Multipurpose Stereo Cameras Based Interactive Event Detection System}
  \Description{a Multi Stereo Cameras based system for scene understanding, event recognition, subject tracking, logging, and notification.}
  \label{fig:pipleine}
\end{teaserfigure}


\maketitle

\section{Introduction}
During the last couple of years, the world has seen a true revolution in terms of media consumption, and media interaction. We went from looking at screens, to interacting with them through buttons. The next revolution came with the touch screen, and since then an infinite number of innovations have come out, Such as, motion tracking for gaming, or behavior analysis for content recommendation. Unfortunately, the impact of this revolution has been limited to the entertainment industry, and has not been able to strongly affect other fields, like industrial applications. This is due to many reasons. One reason is that the technology is not robust enough to be used in a wide range of applications, and is not able to adapt to new environments.

An example of a technology present in the entertainment sector, that has not been able to be widely standardized in other fields \cite{kandroudi2012exploring, pedraza2015rehabilitation}, is motion tracking, which is used for video game console gameplay. The reason for this is that it is lacking in terms of precision when used in large scenes. As a result, except for lab testing, no true attempt has been made to use easily available, commercial stereo vision devices, in other more serious applications like industrial automation, or smart home systems. Therefore, We have found that the proposal of a system that is adaptable for a wide range of use cases from entertainment media consumption, to sensitive decision-making, to real world industrial applications, is a very interesting and challenging problem. We believe that the solution to this problem lies in the combination of multiple technologies, and the ability to adapt them to the environment and the use case.

\section{Related Works}

In this section, we will present some of the work that has already been done regarding the different parts and approaches of the systems that we propose.

\subsection{Stereo Cameras for Event Detection}
We can deduce a number of approaches related to action recognition in both mono and stereo vision \cite{7371223}. It should be noted that most methods apply a combination of these approaches for better results.

\subsubsection{Rule Based :} These methods \cite{s21175895} are based on the use of handcrafted, and hard coded features. They are often used in conjunction with machine learning algorithms. The main advantage of this approach is its speed. The main disadvantage is that it is not very robust, and is not able to adapt to new environments. As we are forced to manually define the logic for the action detection instead of using intelligent automated algorithms.

\subsubsection{Deep Learning :} This approach relies on teaching different types of Deep Neural Networks \cite{jiaxin2021review, veeriah2015differential}, how to understand sequences of video or 3D sensor data, in order to perform action recognition. The main advantage of this approach is that it is very robust and very powerful when it comes to learning complex features, and actions. The main disadvantage is that it is very computationally expensive, and a lot of data is required to train these networks.

\subsubsection{Spatio-temporal Features :} By using different baseline solutions like Spatio-temporal Pyramids \cite{6178760}, we can capture the relationship between the data samples in regard to time and space. Then, we extract features that can be fed into classification models for action recognition.

\subsubsection{Attention Mechanisms :} Through identifying the most relevant parts of the scene using attention \cite{s18071979}, we can reduce the amount of scene data given to the action recognition system. Hence, non-relevant data is discarded. As a consequence, a more precise and efficient training and inference pipeline is guaranteed.

\subsection{Multi Camera Human Tracking and Identification}
Tracking moving objects through different cameras and input sources is a challenging problem. Even though we identify some wildly used approaches, there are other methods that we may not have mentioned in this preliminary review.

\subsubsection{SORT methods :} Operating in real time, SORT methods usually preform object detection, followed by position prediction to predict the trajectories of the objects in the next frame. Variants include SORT \cite{DBLP:journals/corr/BewleyGORU16} and DeepSORT \cite{wojke_bewley_paulus_2017}.

\subsubsection{Siamese Neural Networks :} The idea behind this approach is to teach the network to differentiate between two images, where only one of them contains the tracked object. Through this, the network learns when an object is present in a certain position, and when it's not. During tracking, subsequent frames are compared to find the object throughout a number of consecutive frames \cite{10.1007/978-3-319-48881-3_56, an_yan_2021}.

\section{Proposed Approach}
We propose a system that is robust enough to be adaptable for a wide range of use cases from interactive media content consumption, to real world industrial applications. Using multiple Stereo Cameras, the system captures the scene from multiple angles. Then, reconstructs it into a large 3D Pointcloud \cite{guo_wang_hu_liu_liu_bennamoun_2019}. This process is followed by object detection and tracking, to capture changes in the scene in relation to time and space. Finally, event detection and action recognition are preformed, for overall scene understanding. Figure~\ref{fig:pipleine} represents the proposed pipeline.

Through Scene Understanding, we can achieve an intelligent system that is able to interact with users and be interacted with in a way that is natural and non-disruptive, to the domain of application or the use case.

These possible interactions between subjects in the scene and the system, can either be one-way or two-way. In the case of the former, the system is able to detect and recognize events in the scene, to react to them. For example, in a retail store, the system can detect when a customer is looking at a product and displays the product's information on a screen. Another example is tracking children in a classroom, and analyzing their behavior. Then, giving audible feedback using a friendly voice, as a way to encourage good behavior and discourage bad one. Finally, dangerous action detection is another interesting use case, where workers are notified and actions are logged and reported for safety and insurance purposes.

As for two-way interactions, we are able to add an extra layer of robustness to the pipeline. By giving the subjects the ability to input data into the system, and give feedback on the system's behavior. Hence, we can drastically improve in real time its accuracy, and push for a more cross-domain adaptable solution. In the case of a retail store, the system can detect when a customer is moving towards a certain location, or looking at a certain product. As a consequence, products will be recommended to them through a notification on a wearable device. The customer can interact with the wearable device to give feedback on the recommendation, such as rating it, or adding it to a shopping cart, or ignoring it. This would teach the system to give better recommendations.

\subsection{ZED Stereo Camera}

Recent studies \cite{machines10030183} prove that the ZED stereo vision system out preforms all other similarly priced solutions on the market. We are able to record video at 2.2K (2208x1242), 1080p (1920x1080) and 720p (1280x720). With a depth range between 0.2 and 20 meters. It also comes with useful motion sensors, such as an Accelerometer and a Gyroscope. The fact that it provides an SDK that allows for the extraction of high quality images, depth maps, and point clouds. As well as preform real time processing, like object detection and tracking, makes it a very attractive solution for our use case.

\begin{figure}[h]
  \centering
  \begin{minipage}[b]{\textwidth}
    \includegraphics[width=\linewidth]{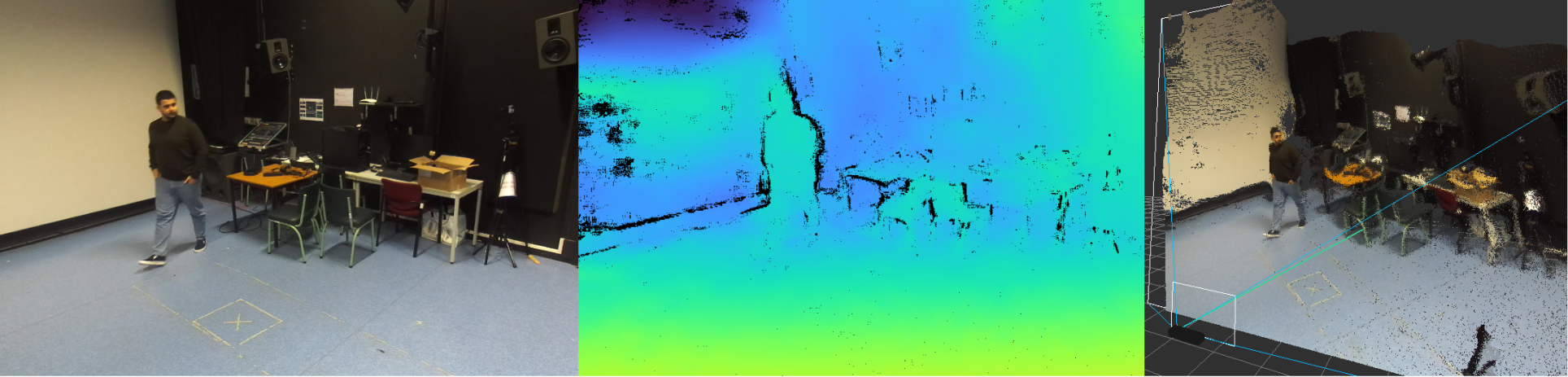}
  \end{minipage}
  \caption{Data acquired through the ZED SDK (left to right: 2D RGB, 2D Depth, 3D Pointcloud)}
  \Description{Example of a scene captured using the ZED, and the data that can be acquired for each frame through the SDK.}
  \label{zedCameraOutputExample}
\end{figure}

\subsection{Object Detection}
\subsubsection{2D Methods :} We look at comparative studies between Deep learning based object detection methods, where we can see that one stage detectors like YOLO and SSD \cite{liu_anguelov_erhan_szegedy_reed_fu_berg_2016} come on top when compared to two stage detectors like Faster R-CNN \cite{ren_he_girshick_sun_2015}. While two stage detectors guarantee higher accuracy over smaller datasets, as well as when detecting smaller objects, One stage detectors allow for real time object detection which is necessary for real time danger detection. A Review of YOLO Algorithm Development \cite{jiang2022review} shows how YOLOv5 is much better than previous versions, in terms of model size flexibility, which translates to higher FPS on edge resources, and accuracy over different model sizes. It also fits better due to its use of different data processing techniques in the network, such as data augmentation through mosaic enhancements. The much newer approaches YOLOv6 and YOLOv7 show very promising results, when compared to other methods. But the YOLOv5 community provides multiple pre-trained models on COCO, with easy to use training and transfer learning scripts, that function on multiple system architectures. A more recent study \cite{mittal2019survey}, on the usage of YOLO on Edge devices, compares inference speed across YOLO versions on the Jetson AGX Xavier. It confirms that YOLOv5 provides a good balance between FPS and accuracy. A more recent, more powerful YOLOv8 was recently made available, and it is an approach that we need to test.

\subsubsection{From 2D to 3D :} Through the high quality depth maps generated by the ZED, we are able to project the 2D bounding boxes detected on the left or the right camera outputs to 3D space $(X, Y, Z)$ using the following equations :

\begin{equation}
  Z = \text{Depth value of pixel (u, v) from depth map}
\end{equation}
\begin{equation}
  X = \frac{ (u - c_{x}) \times Z }{f_{x}}
\end{equation}
\begin{equation}
  Y = \frac{ (v - c_{y}) \times Z }{f_{y}}
\end{equation}

$(u, v)$ is the pixel position in the image and the depth map. $(f_{x}, f_{y})$ are the focal length \cite{doi:10.1080/10407413.2014.877284} in pixel units of the camera. $(c_{x}, c_{y})$ coordinates of the principal point of the camera in pixel units. These values can be extracted from live video feed or recordings using the ZED SDK.

\subsection{Event Recognition}
We have only tested with primitive hard-coded rule based event recognition. By calculating the distances between the detected bounding boxes, we can establish relationships between objects. Hence, recognize events. A Person, getting closer each frame to a turned-ON light switch, may indicate the action of switching the light OFF. More advanced event recognition techniques may be added in future implementations.

\section{Initial Experimentation}

\subsection{ZED Camera Accuracy Testing}

\begin{figure}[h]
  \centering
  \begin{minipage}[b]{0.95\textwidth}
    \includegraphics[width=\linewidth]{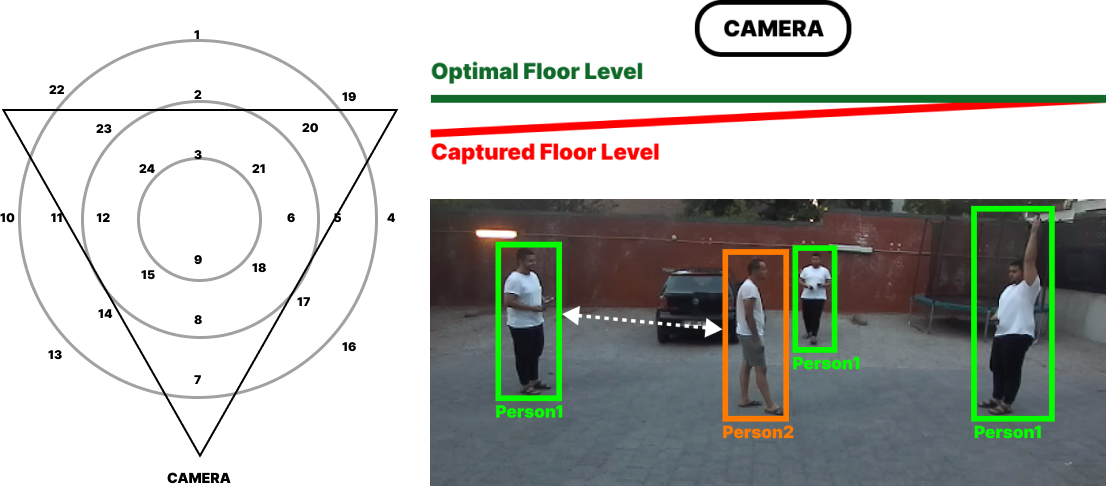}
  \end{minipage}
  \caption{Setup used for testing the ZED Camera's Accuracy.}
  \Description{Left image shows the poisiton from which real life distance measurements were taken, Top left image is the camera level from the ground and Bottom Left is an example of the results.}
  \label{precisionTestingSetup}
\end{figure}

To insure quality and provide a reliable solution, we opted for testing the ZED Camera's capture capabilities and depth estimation precision in different lighting conditions. As seen in Figure~\ref{precisionTestingSetup}, We positioned a subject "Person2" in the middle of the scene, and had another subject "Person1" Capture real life distance measurements to "Person2", using a laser distance measuring device, at different positions, circling "Person2". We have provided the positions from which the measurements were taken. The captured floor level isn't perfectly leveled, and this was done on purpose to study the effect of a non-levelled environment, on the depth sensing capabilities of the camera. The furthest point of measurement from the camera is 12.5 meters. The distance from each diagonal position to the one before or after it is closely similar. The same can be said for horizontal and vertical positions.

\begin{figure}[h]
  \centering
  \begin{minipage}[b]{0.95\textwidth}
    \begin{minipage}[b]{0.23\textwidth}
      \centering
      \includegraphics[width=\linewidth]{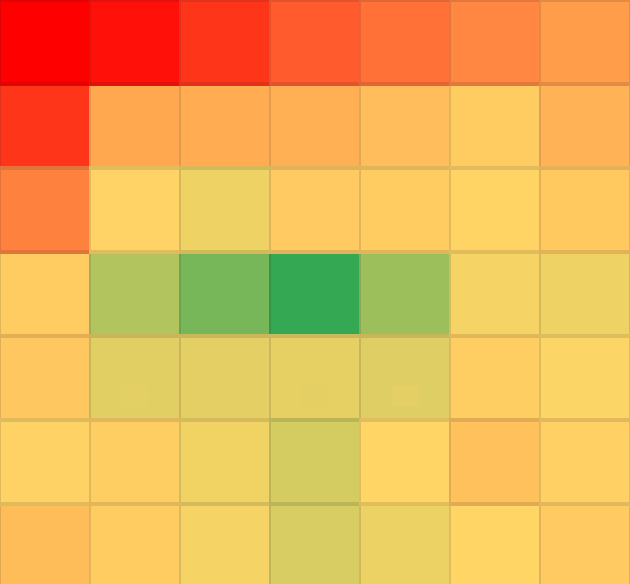}
      \subcaption{720P Morning}
    \end{minipage}%
    \hfill
    \begin{minipage}[b]{0.23\textwidth}
      \centering
      \includegraphics[width=\linewidth]{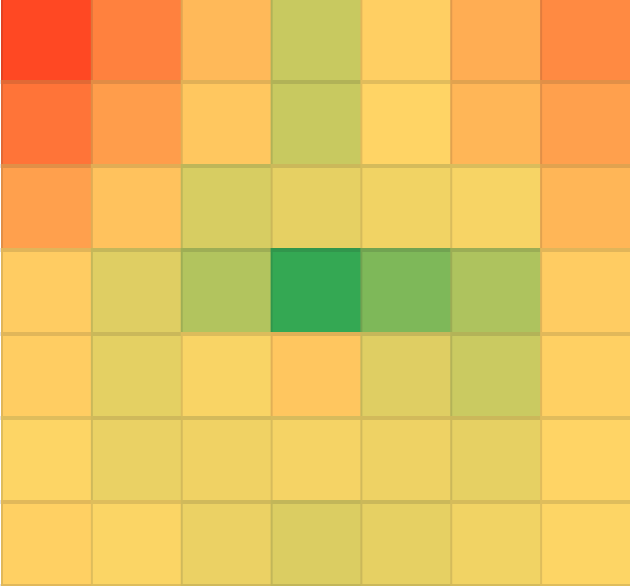}
      \subcaption{2K Morning}
    \end{minipage}%
    \hfill
    \begin{minipage}[b]{0.23\textwidth}
      \centering
      \includegraphics[width=\linewidth]{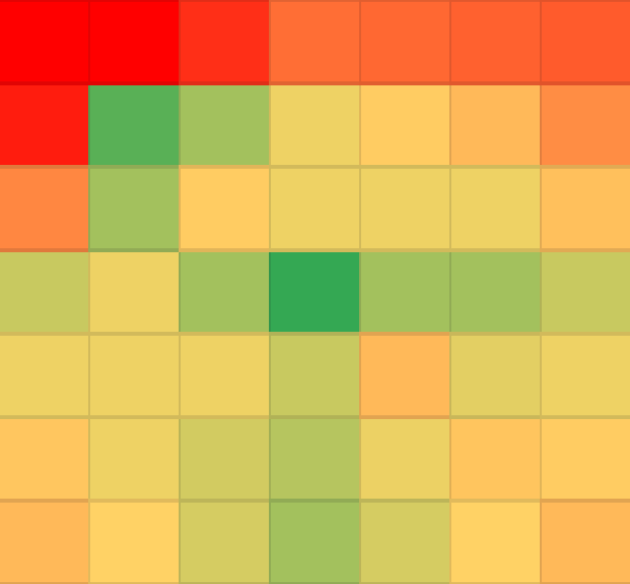}
      \subcaption{2K Night}
    \end{minipage}%
    \hfill
    \begin{minipage}[b]{0.23\textwidth}
      \centering
      \includegraphics[width=\linewidth]{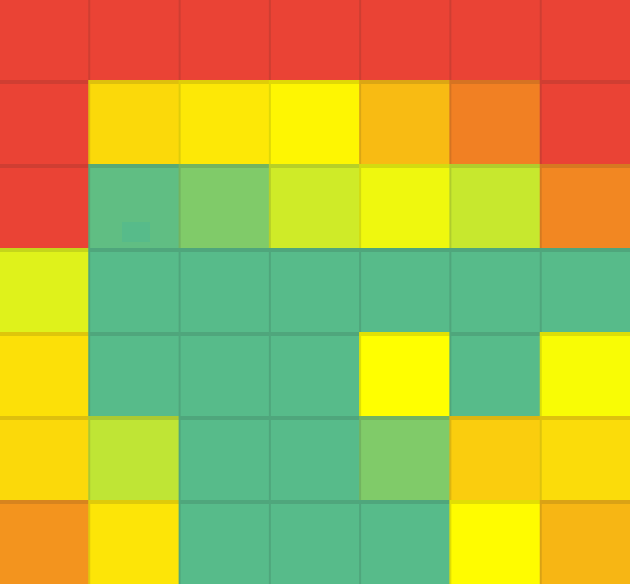}
      \subcaption{Average}
    \end{minipage}%
  \end{minipage}%

  \caption{Preliminary Results of the ZED Camera Accuracy Testing}
  \Description{All heatmaps correspond to the positions used during testing. The average heatmap is calculated through averaging the results for all heatmaps and increasing saturation. Each subcaption corresponds to the camera resolution used to capture the scene, as well as the time of the day.}
\end{figure}

A YOLOv5n (4.5 FLOPs(B)) model pretrained on the COCO dataset is used to detect the subjects' positions in the scene. Depth maps are used to project the bounding boxes to 3D, and distances between them are calculated. The distance measurement results are compared to the real life measurements, and heatmaps are produced showing the differences between the two types of measurements. Over 1 meter of error is red, while measurements under 30 centimeters of error are green, everything in between is automatically colored in that color range.

\subsection{Data Collection and Initial Use Case}

In order to have direct access to a possible user base for the system, and be able to test its full potential in a real production environment, we chose to collaborate with a Belgian railway infrastructure construction company to implement the proposed pipeline, for collision detection, and zoning regulation. The company has many construction sites, and we were able to collect data and run some tests on some of them.

\begin{figure}[h]
  \centering
  \includegraphics[width=\linewidth]{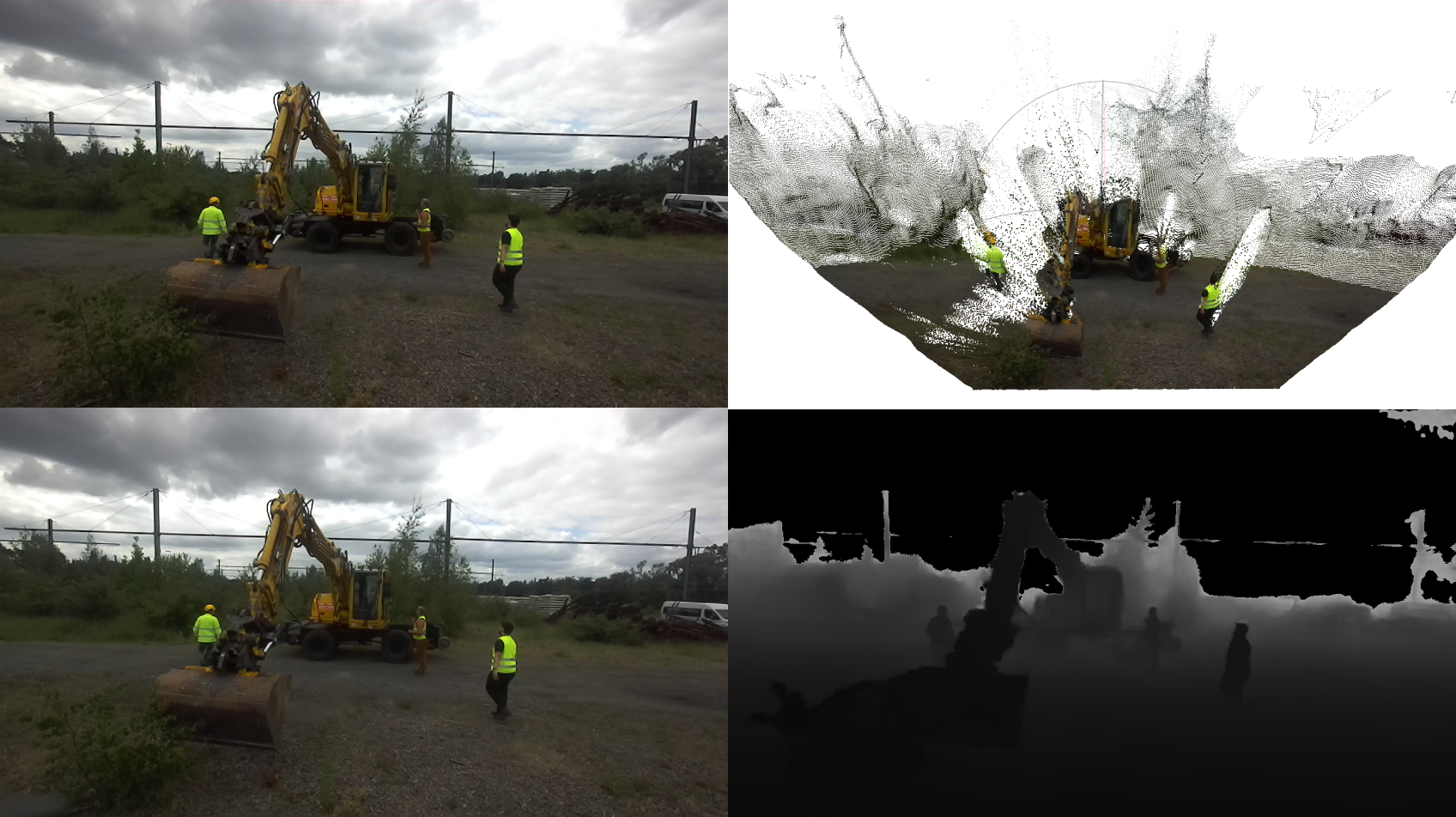}

  \caption{Example of ZED output: left camera view (top left), right camera view (bottom left), estimated depth map (bottom right), reconstructed 3D pointcloud (top right).}
  \Description{Different outputs of the ZED Camera View from construction site recording}
  \label{fig:zed-output}
\end{figure}

Initially, We only recorded data at 720P resolution. Using the ZED's SDK we extracted left and right images, a depth map, as well as a point cloud for each frame as seen in Figure~\ref{fig:zed-output}. This dataset was partially annotated by railway experts, for event detection and action recognition future research. We have also annotated it for excavator parts object detection and tracking.

We recorded in three separate construction sites, for 1 to 2 hours each time. Two of the sites had rails, an excavator, and some workers. The third site was mainly to test an object detection model that we created for excavator parts detection. Therefore, we only had an excavator with a number of construction workers preforming dangerous actions.

\subsection{Zoning, Tracking and Feedback}

Zone creation and management is just an example of a possible application for the pipeline. We have tested it in a construction site where it's necessary to define a specific zone for the workers to stay in, and notify them with an audio alarm when they leave. Movement in the zone is also tracked to insure the safety of the workers interacting directly with the excavator arm.

\begin{figure}[h]
  \centering
  \begin{minipage}[b]{\textwidth}
    \begin{minipage}[b]{0.75\textwidth}
      \centering
      \includegraphics[width=\linewidth]{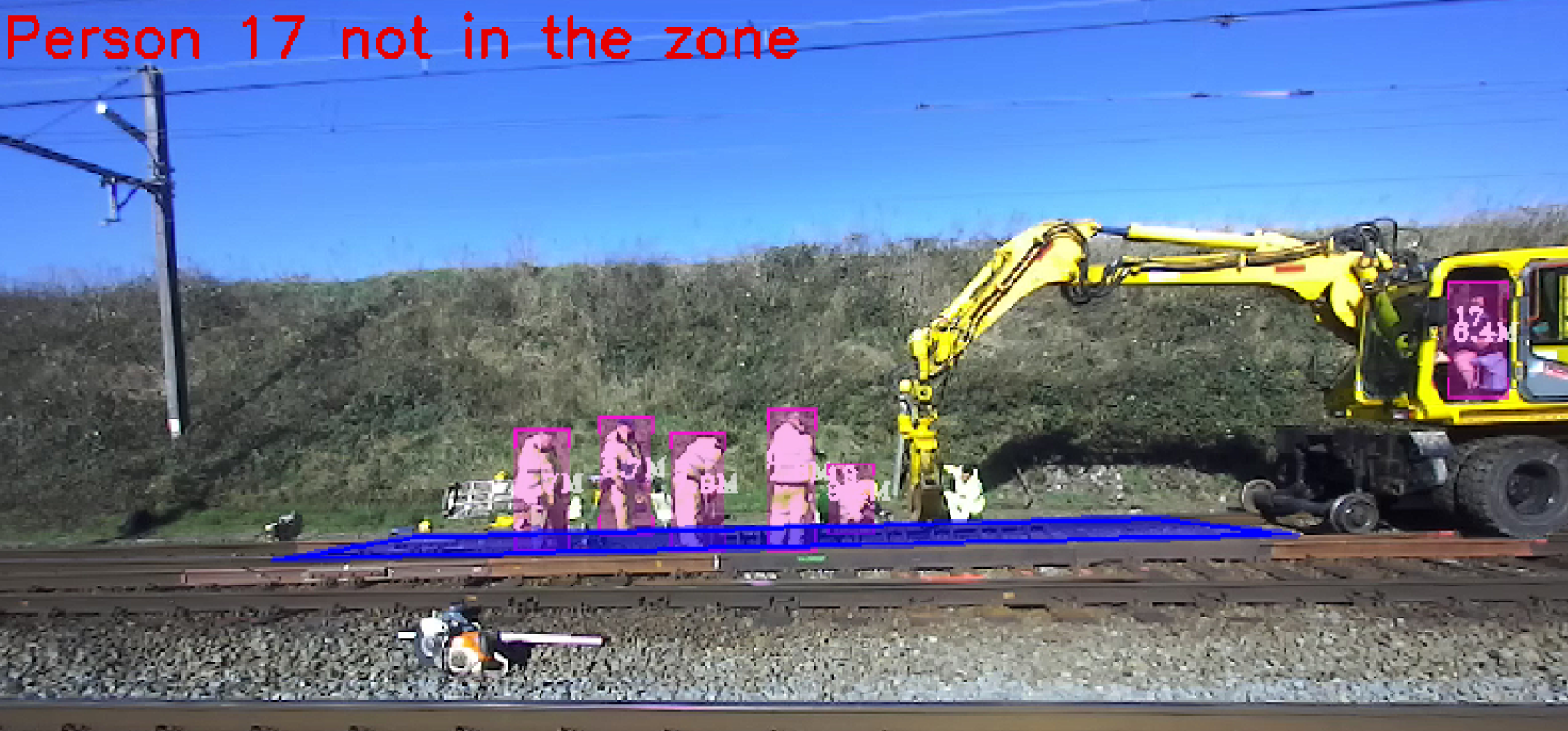}
      \subcaption{Font View}
    \end{minipage}%
    \hfill
    \begin{minipage}[b]{0.208\textwidth}
      \centering
      \includegraphics[width=\linewidth]{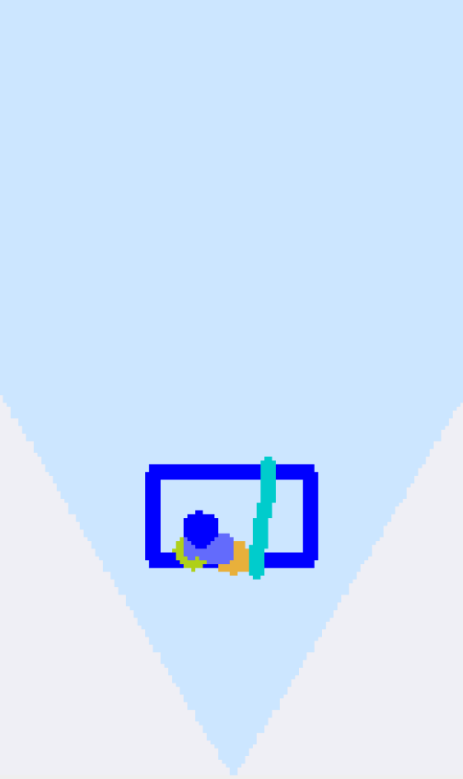}
      \subcaption{Top Down View}
    \end{minipage}%
  \end{minipage}%

  \caption{Zoning in a Construction Site with Logging and Notifications}
  \Description{The Zones can be defined in 2D, and then projected in 3D. The top down view shows the zones in 3D, and the font view shows the zones in 2D.}
\end{figure}

\subsection{Pipeline Generalization and Other Use Cases}

We have also experimented with filming classrooms for tracking student behavior, but we haven't explored that use case yet. Another test that we have done was zoning in a smart home to turn on and off devices automatically. For example, we can turn on a light, when someone gets close to the area that the light covers. The same was tested with Music playing, by increasing sound the further we get from a speaker.

\begin{figure}[h]
  \centering
  \includegraphics[width=\linewidth]{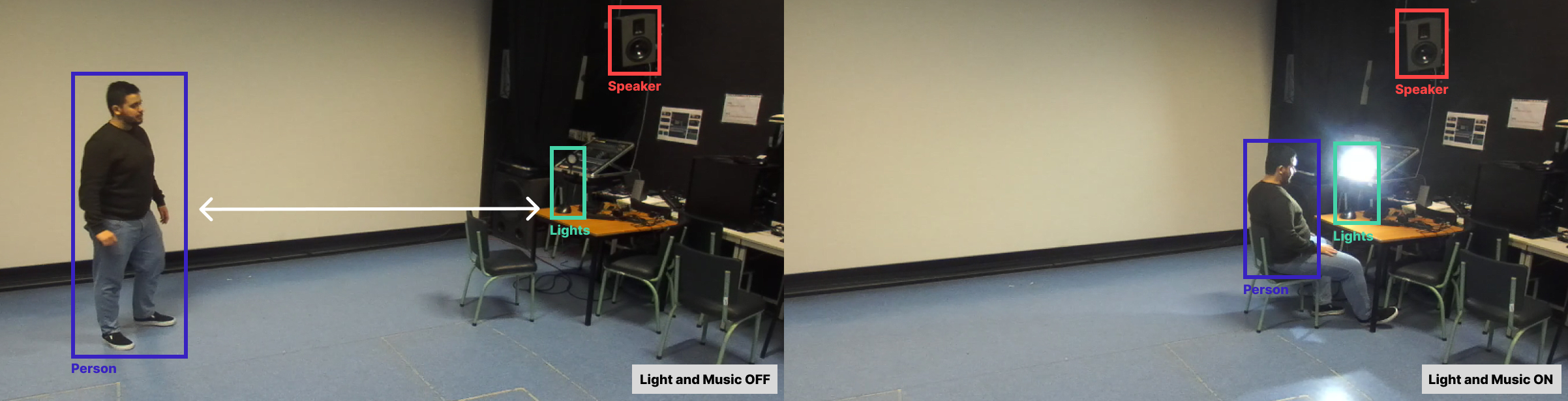}

  \caption{Distance and Zone Based Event Triggering (Light Switching and Audio Feedback)}
  \Description{We can see a person getting close to detected objects (light and speakers) and the objects are turned on or off depending on the distance of the person to those objects.}
  \label{fig:musicLightTesting}
\end{figure}

\begin{figure}[h]
  \centering
  \includegraphics[width=\linewidth]{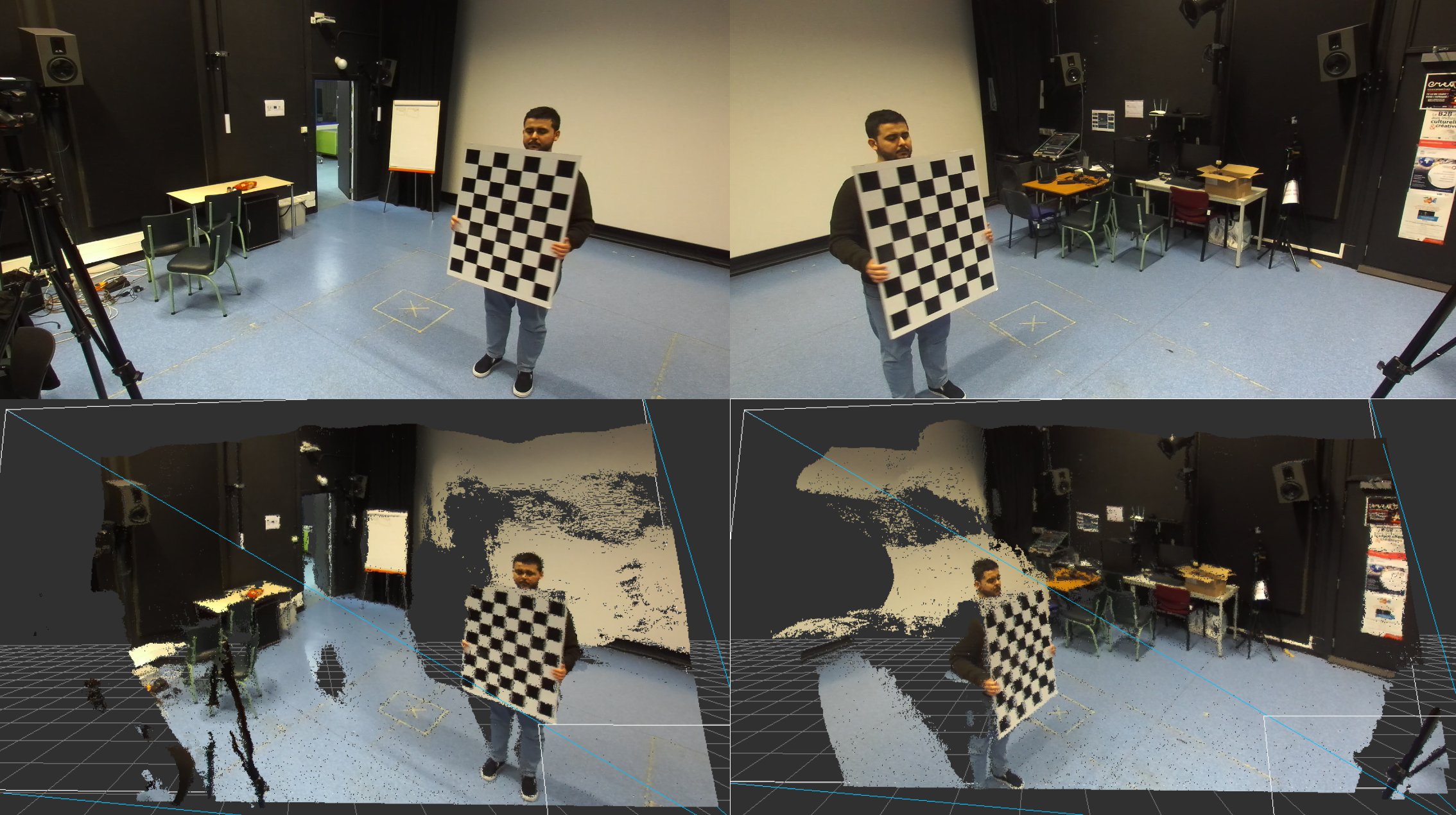}

  \caption{Checkerboard Based Multi-Camera Calibration Testing for Pointcloud Registration}
  \Description{A Person Holding a checkboard pattern for callibation of multiple cameras}
  \label{fig:FusionMultipleCamerasCallibration}
\end{figure}

In some cases, generalizing the pipeline requires the usage of multiple cameras. In the case where multiple rooms exist, we fuse multiple pointclouds from each room, to have a full view of the environment. The fused pointcloud can be used to track residents, define more accurate zones, and have better scene understanding. For that, we experimented with adding multi-camera calibration solutions to the pipeline, As seen in Figure~\ref{fig:FusionMultipleCamerasCallibration}. We  recorded multiple views of the same scene with a checkerboard pattern, and we are experimenting with End-to-End Pointcloud Registration, Voxel Grid Fusion, and other approaches.

\subsection{Deployment and Processing}

For testing the solutions' ability to adapt to different use cases, and to insure its generalization, we opted for testing on different types of devices. The ZED Camera has a minimum requirement of 4GB RAM, and an NVIDIA GPU with compute capability higher than 3.0. For centralized systems like Smart Homes, Indoor Gaming Systems, Video Consumption, and general traditional interactive experiences, we tested on a desktop computer. For a more mobile and portable solution that can be used in production environments, serious pipelines, and On-Terrain applications, we tested on a Rugged Windows Tablet. For a more Up-to-date compact deployment solution, we tested on NVIDIA edge resources. The specifications of the devices used for testing can be seen in Table~\ref{tab:devicesTestingList}.

During testing, each pass through the pipeline was measured as a single frame. Table~\ref{tab:devicesTestingResults} shows the FPS value for capturing from the ZED, detecting objects, and processing the data to preform different lightweight operations, like audio notification or object position logging. The results from different tests have been averaged and combined into a single table. We test with both live input from the camera, and prerecorded SVO files, to represent how different factors like Disk I/O speed, and USB transfer speed affect performance.

\begin{table}
  \caption{Devices Used for Deployment and Testings of the Proposed Pipeline}
  \label{tab:devicesTestingList}
  \begin{tabular}{@{}cccccc@{}}
    \toprule
    Device                                                                            & CPU                                                                                        & GPU                                                                         & Memory                                                                         & OS           & Examples of Usage                                                                          \\ \midrule
    \begin{tabular}[c]{@{}c@{}}[EDGE]\\ NVIDIA Jetson Orin\end{tabular}               & \begin{tabular}[c]{@{}c@{}}12-core A78AE \\ Armv8.2 64-bit \\ 3MB L2\\ 6MB L3\end{tabular} & \begin{tabular}[c]{@{}c@{}}2048-core Ampere \\ 64 Tensor Cores\end{tabular} & \begin{tabular}[c]{@{}c@{}}64GB \\ 256-bit \\ LPDDR5 \\ 204.8GB/s\end{tabular} & Jetpack 5.1  & \begin{tabular}[c]{@{}c@{}}Embedded Systems\\ Distributed Systems \\Compact Solutions\end{tabular}             \\ \addlinespace[0.5em]
    \begin{tabular}[c]{@{}c@{}}[EDGE]\\ NVIDIA Jetson Xavier\end{tabular}             & \begin{tabular}[c]{@{}c@{}}8-core Carmel \\ Armv8.2 64-bit \\ 8MB L2\\ 4MB L3\end{tabular} & \begin{tabular}[c]{@{}c@{}}512-core Volta  \\ 64 Tensor cores\end{tabular}  & \begin{tabular}[c]{@{}c@{}}32GB \\ 256-bit \\ LPDDR4x\\ 136.5GB/s\end{tabular} & Jetpack 4.6  & \begin{tabular}[c]{@{}c@{}}Embedded Systems\\ Distributed Systems \\Compact Solutions\end{tabular}             \\ \addlinespace[0.5em]
    \begin{tabular}[c]{@{}c@{}}[LAPTOP]\\ Lenovo T15g Gen2\end{tabular}               & \begin{tabular}[c]{@{}c@{}}Intel© \\ i7-11850H \\ 8 Cores\\ x2.5GHz\end{tabular}           & \begin{tabular}[c]{@{}c@{}}GeForce \\ RTX 3070 \\ 8GB Max-Q\end{tabular}    & \begin{tabular}[c]{@{}c@{}}16GB \\ DDR4\\ 3200MHz\end{tabular}                 & Ubuntu 20.04 & \begin{tabular}[c]{@{}c@{}}Smart Home Servers\\ Workstations\\ Gaming Systems\end{tabular} \\ \addlinespace[0.5em]
    \begin{tabular}[c]{@{}c@{}}[TABLET]\\ DT340T RUGGED \\ 2-IN-1 TABLET\end{tabular} & \begin{tabular}[c]{@{}c@{}}Intel® \\ i5-8250U \\ 4 Cores \\ x1.6GHz\end{tabular}           & \begin{tabular}[c]{@{}c@{}}GeForce \\ GTX 1050 \\ 4GB\end{tabular}          & \begin{tabular}[c]{@{}c@{}}8GB \\ DDR4\\ 2400MHz\end{tabular}                  & Windows 10   & \begin{tabular}[c]{@{}c@{}}Portable Solutions\\ On-Site Testing\end{tabular}               \\ \bottomrule
  \end{tabular}
\end{table}

\begin{table}
  \caption{FPS results for testing on different devices}
  \label{tab:devicesTestingResults}
  \begin{tabular}{@{}ccccc@{}}
    \toprule
    Device & Single Camera & Two Cameras & 1 File & 2 Files \\ \midrule
    Orin   & 30            & 14          & 18     & 8       \\
    Xavier & 11            & /           & 10     & 4       \\
    Laptop & 20             & 10           & 15      & 8       \\
    Tablet & 8             & /           & 9      & 3       \\ \bottomrule
  \end{tabular}
\end{table}

\section{Future Work}

\subsection{Accuracy Testing}
What was done in terms of Accuracy testing is not sufficient to understand how the ZED's depth estimation accuracy effects the system. Testing in different lighting conditions, different floor levels, different camera angles or orientations, and different camera positions is required.

\subsection{Multi-Camera Fusion}
A single camera is not enough to cover all the angles and positions in extremely complex scenes. We need to further test and implement multi-camera fusion to generate multiple points clouds, fuse them, and get a bigger and more accurate view of the captured environment. This is necessary to avoid occlusion problems, and have better scene understanding. In the future, we will be using a mix of registration and voxelization \cite{rs11010092}. By doing an initial pointcloud registration, we should get a fairly inaccurate fused point cloud, but still be able to have some matching points. Then, we can use voxelization to simplify the fused point cloud and reduce registration errors, as well as reduce pointcloud size by decreasing the number of points.

\subsection{Object Detection}
Even though, we tested with the ZED SDK's Object Detection Model, and we trained a custom YOLOv5n (4.5 FLOPs(B)) model for the case of detecting excavators in construction sites. We want to test multiple YOLO versions, and other object detection models like SSD, and Faster R-CNN. Comparing the performance of many object detection methods on the Jetson Xavier and Orin would be a significant contribution, From another point, Using voxelization for multi-camera fusion, allows us to explore the usage of purely synthetic data for training. By simplifying the pointcloud to a voxelized shape, there is a possibility that the models can be trained through procedurally generated synthetic voxelized objects. This can push for a more easily adaptable and robust system.
\subsection{Domain Adaptation and Pipeline Optimization}

Optimization is necessary if we want to have a system that is easily adaptable, and very compatible with as many devices as possible. We will explore the usage of Pruning and Quantization before model deployment. In addition, we have already explored the usage of Data Knowledge Distillation to have the system adapt to new environments, by retraining less architecturally complex models using very small amounts of data. These models can be quite performant and accurate, even with new object classes.

\subsection{User Feedback and Input}
We have discussed many aspects of the system in regard to input, logging, and tracking. But we haven't addressed the output, notification, and feedback aspects, as much as we would like to. It will be interesting to test different approaches. By taking into account different factors like ease of use, implementation steps, cost, etc. We should define rules for the decision-making process in regard to what device to use in which situation.

\subsubsection{One way communication devices :} We can use different types of devices, either visual or audio, to notify the user that an event is happening, like playing music, or that they are inside a zone, like in the case of construction sites.

\subsubsection{Two way communication devices :} Giving the user the ability to interact with the system is crucial for interactivity. Smartwatches are a good start, but we can also explore the usage of smart glasses, or even smart rings. These devices need too much commitment from the user, when it comes to wearing them in daily life. But they make more sense in specific interactive experiences like gaming. Other options can be used in cases where users should not be given a choice, when it comes to wearing the device. In the case of construction sites, smart helmets are not a very good option. Workers tend to remove them, as falling objects are rare in comparison to dangers coming from stepping on a nail, or a sharp object. Therefore, the usage of smart shoes or smart vests is more appropriate.

\begin{table}[h]
  \caption{Some Available Feedback Devices (Input : ability to send data to the system, Output: ability to notify the user)}
  \label{tab:possibleInputOutputDevices}
  \begin{tabular}{@{}cccccc@{}}
  \toprule
  Device & Type & Input & Ouput & ~ Pricing (€) & Developer Access \\ \midrule
  Apple Watch Series 8 & Watch & \cmark & \cmark & 499 & \cmark \\
  Galaxy Watch 5 & Watch & \cmark & \cmark & 239 & \cmark \\
  Oura & Ring & \cmark & \xmark & 314 + 5.99/mo & \cmark \\
  Digitsole 2.0 & Shoe & \cmark & \xmark & 180 & \xmark \\
  Hexoskin Pro Kit & Shirt & \cmark & \xmark & 648 & \cmark \\
  Vuzix Blade 2 & Glasses & \cmark & \cmark & 1699 & \cmark
  \end{tabular}
\end{table}

\section{Conclusion}

The pipeline that we propose is very large in the way that it opens many questions to be solved, and challenges to be tackled. Object detection, multi camera fusion, optimization, deployment, domain adaptability, and user understanding through notification and feedback. We may seem to be tackling a lot of different challenges. But what we have tried to do in this work was not providing a complete working system, but present the baseline that we have developed, and explain the initial experimentation in both lab environment testing and real world testing. Finally, we outline the strategy we intend to pursue in order to develop a more comprehensive and deployable system.





\bibliographystyle{ACM-Reference-Format}
\bibliography{sample-base}

\end{document}